\documentclass{article}

\usepackage{PRIMEarxiv}

\usepackage[utf8]{inputenc} 
\usepackage[T1]{fontenc}    
\usepackage{hyperref}       
\usepackage{url}            
\usepackage{booktabs}       
\usepackage{amsfonts}       
\usepackage{nicefrac}       
\usepackage{microtype}      
\usepackage{lipsum}
\usepackage{fancyhdr}       
\usepackage{graphicx}       
\graphicspath{{media/}}     

\usepackage{tikz}
\usepackage{epigraph}
\usepackage[shortlabels]{enumitem}
\usepackage{xcolor,colortbl}
\usepackage{array, makecell, multirow, tabularx}
\usepackage{multicol}
\usepackage{threeparttable}
\usepackage{rotating}
\usepackage{nicematrix}
\usepackage{adjustbox}
\usepackage{framed}
\usepackage{color,soul}
\usepackage[most]{tcolorbox}
\usepackage{subcaption} 
\usepackage{listings}
\usepackage{academicons}
\usepackage{enumitem}
\usepackage{lipsum}
\usepackage{float}
\usepackage{orcidlink}

\lstdefinestyle{myStyle}{
    belowcaptionskip=1\baselineskip,
    breaklines=true,
    frame=none,
    numbers=none,
    basicstyle=\footnotesize\ttfamily,
    keywordstyle=\bfseries\color{green!40!black},
    commentstyle=\itshape\color{purple!40!black},
    identifierstyle=\color{blue},
    backgroundcolor=\color{gray!10!white},
}
\definecolor{codegreen}{rgb}{0,0.6,0}
\definecolor{codegray}{rgb}{0.5,0.5,0.5}
\definecolor{codepurple}{rgb}{0.5,0,0.82}
\definecolor{backcolour}{rgb}{0.95,0.95,0.92}

\usepackage{capt-of}
\usepackage{verbatim}
\usepackage{outlines}
\usepackage{cleveref}
\usepackage[numbers]{natbib}
\usepackage{rotating}

\usepackage[group-separator={,}]{siunitx}
\title{Improving Existing Optimization Algorithms with LLMs
}

\author{
  Camilo Chacón Sartori \orcidlink{0000-0002-8543-9893}\thanks{Corresponding author: \texttt{cchacon@iiia.csic.es} }\\
  Artificial Intelligence Research Institute (IIIA-CSIC)\\
  Bellaterra, Spain\\
  \texttt{cchacon@iiia.csic.es}
  \and
  Christian Blum \orcidlink{0000-0002-1736-3559}\\
  Artificial Intelligence Research Institute (IIIA-CSIC)\\
  Bellaterra, Spain\\
  \texttt{christian.blum@iiia.csic.es}
}

\begin{document}
\maketitle

\begin{figure}[h]
\centering
  \includegraphics[width=\linewidth]{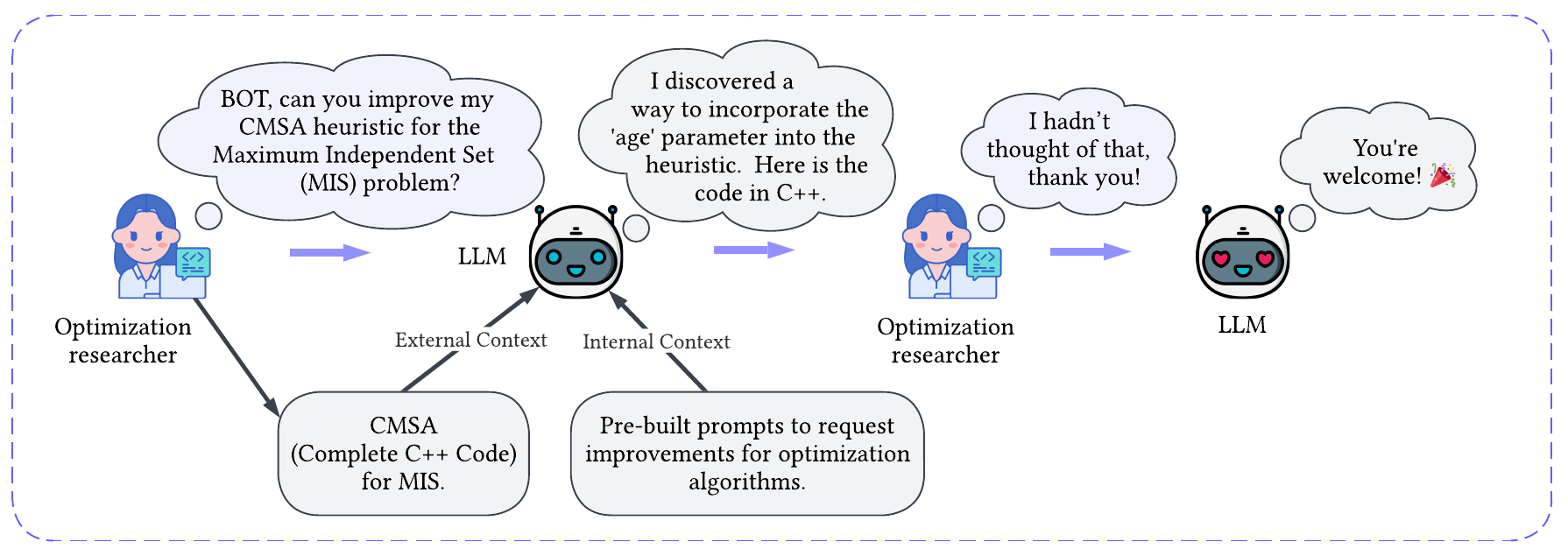}
  \caption{A dialogue showing how a chatbot applies our approach to improving optimization algorithms.}
  \label{fig:teaser}
\end{figure}

\begin{abstract}
The integration of Large Language Models (LLMs) into optimization has created a powerful synergy, opening exciting research opportunities. This paper investigates how LLMs can enhance existing optimization algorithms. Using their pre-trained knowledge, we demonstrate their ability to propose innovative heuristic variations and implementation strategies. To evaluate this, we applied a non-trivial optimization algorithm, Construct, Merge, Solve \& Adapt (CMSA)---a hybrid metaheuristic for combinatorial optimization problems that incorporates a heuristic in the solution construction phase. Our results show that an alternative heuristic proposed by GPT-4o outperforms the expert-designed heuristic of CMSA, with the performance gap widening on larger and denser graphs. Project URL: \url{https://imp-opt-algo-llms.surge.sh/}
\end{abstract}


\section{Introduction}
\epigraph{\hfill An algorithm must be seen to be believed.}{--- \textup{Donald Knuth}}
There is a wide variety of optimization algorithms of all kinds and flavors. A simple search using the term `optimization algorithm' in databases like Scopus or platforms like GitHub yields thousands of results, with that number growing every year. Additionally, optimization researchers often maintain private collections of their algorithms. All these algorithms---open-source or proprietary, and implemented in diverse programming languages---can potentially be improved. By refining their original code with modern techniques and technologies, we can achieve more efficient designs and implementations that go beyond what their creators originally envisioned.

In recent years, the development and growth of Large Language Models (LLMs)---popularized by models such as OpenAI’s GPT-4~\cite{openai2024gpt4technicalreport}, Anthropic’s Claude~\cite{anthropicIntroducingClaude}, Google’s Gemini~\cite{geminiteam2024gemini15unlockingmultimodal}, Meta's Llama 3~\cite{grattafiori2024llama3herdmodels}, and recently DeepSeek~\cite{deepseekai2024deepseekv3technicalreport}---has opened the door to a wealth of new possibilities. Among the most transformative advancements is code generation. Tools like GitHub Copilot\footnote{\url{https://github.com/features/copilot}}, which integrates seamlessly with code editors like Visual Studio Code, and Cursor\footnote{\url{https://www.cursor.com/}}, an editor with built-in LLM capabilities, have become essential tools for many software developers, revolutionizing their daily workflows. For example, a Python developer could request an LLM to generate a template for invoking external APIs, and the LLM would automatically produce the required code. LLMs have become invaluable for streamlining coding tasks, especially those that are routine and highly repetitive~\cite{Jiang2024-io}. Therefore, it is natural to wonder: if LLMs excel at simple programming tasks, could they also aid in improving sophisticated optimization algorithms?

One of the recent examples of leveraging LLMs in optimization algorithms is the development of frameworks for generating new black-box metaheuristics~\cite{vanstein2024llamealargelanguagemodel}. Finding an effective metaheuristic, for example, for a combinatorial optimization problem can be a significant challenge. However, using LLMs to build upon an existing metaheuristic as context and employing them as tools to discover new heuristics or more efficient implementations (e.g., reducing RAM usage or computation time) in a given programming language remains a largely unexplored area. Implementing metaheuristics, unlike other algorithms, requires a focus on computational efficiency, mathematical expressions in scientific computing, and efficient data structure selection. Thus, designing new algorithms in this field demands expert implementation skills. 

In this paper, we demonstrate how LLMs like GPT-4o can be leveraged to enhance a sophisticated optimization algorithm, specifically the Construct, Merge, Solve \& Adapt (CMSA) hybrid metaheuristic~\cite{blum2024construct, BLUM201675, BLUM20114135}. Starting with an expert-developed C++ implementation of CMSA for the Maximum Independent Set (MIS) problem (of approximately 400 code lines), we employed an in-context prompting strategy combined with interactive dialogue. Our results show that the LLM successfully comprehended the CMSA implementation for MIS and discovered novel heuristics while suggesting improvements to the C++ codebase. This successful example opens up new possibilities for enhancing existing complex optimization algorithms by using LLMs as assistants.

The paper unfolds as follows. In Section~\ref{sect:background}, we introduce code generation using LLMs, making it accessible for readers without prior experience in this field. We also explain the MIS problem and provide a brief overview of the CMSA algorithm. Next, in Section~\ref{sec:integration}, we present our methodology for enhancing CMSA for the MIS problem using LLMs through a novel narrative that describes our process and provides detailed steps for result reproduction. Section~\ref{sec:evaluation} presents our experimental results and their interpretation. The limitations of our approach and future research directions are discussed in Section~\ref{sec:disc}. The paper concludes by summarizing our key findings and emphasizing the potential impact of LLMs on existing metaheuristics.

\section{Background}\label{sect:background}
\subsection{Code Generation with LLMs}

Code generation is one of the primary research areas in LLMs~\cite{Jiang2024-io, joel2024surveyllmbasedcodegeneration, chen2024surveyevaluatinglargelanguage}. The concept is straightforward: given a prompt, such as \textit{``I need an algorithm to sort a list of numbers in Python''}, the LLM is expected to return a corresponding algorithm (e.g., \texttt{Quicksort}) implemented in Python. This capability arises because LLMs are trained on massive amounts of data, which include code from diverse programming languages available on the internet. These codes are sourced not only from GitHub repositories but also from StackOverflow, technical documentation, scientific articles, and publicly available books. 

A more explicit prompt can lead to more sophisticated responses. For example, the prompt \textit{``I need an efficient sorting algorithm for a list of numbers in Python, which can be processed in parallel, utilizing modern optimization techniques''} in GPT-4o generates a \texttt{ParallelMergeSort}. While efficient, it can be further refined by interacting with the model. For instance, asking \textit{``Find new ways to improve it''} results in an enhanced version using techniques like optimized \texttt{Quicksort} for small arrays, better memory management with \texttt{numpy}, and heap merge for combining sorted arrays. This approach leverages \texttt{MergeSort}'s ease of parallelization. However, the model might have suggested a different algorithm if the focus was on memory optimization without parallelism. This emphasizes the importance of prompt design and iterative refinement of responses~\cite{zhou2023largelanguagemodelshumanlevel}.

LLM response quality depends on the data used for training, and with the refinement of datasets and the increase in size, their performance in code generation has improved~\cite{zan2023largelanguagemodelsmeet}. Moreover, no LLM is flawless when it comes to generating error-free code. Continuing with the \texttt{ParallelMergeSort} example, a model might return code with bugs, so interaction with the model (e.g., copying and pasting error messages from the Python interpreter) is necessary to refine the responses. One could even ask the model to generate its own test suite to verify its algorithm. Different LLMs for these tasks are generally evaluated through benchmarks. One of the most widely used is \textsc{HumanEval}~\cite{chen2021evaluatinglargelanguagemodels}, which features programming challenges that assess language comprehension, algorithmic competencies, and basic mathematics---some of which are comparable to straightforward software developer interview questions.

Furthermore, \textit{code generation using LLMs} is a broad area, as code may originate or be destined for very different domains, including 
 the following: data science code for analyzing data and building predictive models~\cite{wen2024groundingdatasciencecode}; systems code for managing hardware and low-level operations~\cite{Joel2024-nw, Cummins2023-xg}; frontend code development for web applications and UI~\cite{xiao2024prototype2codeendtoendfrontendcode}; and optimization code for solving complex computational problems. Each domain presents its own unique challenges and requirements, with our focus being on the latter, specifically in the field of metaheuristics. 

Concerning the automatic generation of metaheuristics with LLMs, given the famous ``No Free Lunch Theorem''~\cite{585893}, researchers understand that no single metaheuristic algorithm can outperform all others across all optimization problems. This fundamental principle naturally leads to an interesting possibility: LLMs could serve as powerful automatic generators of black-box metaheuristics, significantly reducing the time needed to find the `best' implementation of a metaheuristic for a specific problem. In this context, LLMs could be tasked with discovering novel variations and operators that create new metaheuristics---potentially even surpassing state-of-the-art algorithms for particular problems~\cite{vanstein2024llamealargelanguagemodel}. 

The first notable demonstration of heuristic discovery using LLMs was conducted by FunSearch~\cite{RomeraParedes2023MathematicalDF}, which showcased the potential of leveraging LLMs to generate novel heuristics for the Bin Packing (BP) Problem through a heuristic evolution process. However, a recent study~\cite{sim2025hypebenchmarkingllmevolvedheuristics} points out that the heuristics developed by FunSearch face challenges in generalizing across diverse BP problem instances. While FunSearch relies on an incomplete or suboptimal base program as a starting point, our approach starts with a complete and carefully designed implementation of an optimization algorithm. Other related works can be summarized as follows.~\citet{vanstein2024llamealargelanguagemodel}'s LlaMEA is a framework that integrates evolutionary algorithms with LLMs to iteratively generate and refine novel black-box metaheuristics during runtime. Similarly, \citet{hemberg2024evolvingcodelargelanguage} proposed LLM GP, combining genetic programming with LLMs to evolve operators through LLM assistance. Meanwhile, \citet{10.1145/3583133.3596401} demonstrated via prompt engineering that LLMs can generate innovative metaheuristics by identifying and decomposing six high-performing swarm algorithms for continuous optimization.

While these studies already give a glimpse of possible use cases for LLMs in optimization, they primarily focus on creating new algorithms rather than improving existing ones---that is, building upon pre-existing complex code. This distinction is significant given the vast landscape of published optimization algorithms---a simple GitHub search for `optimization algorithm' returns over 20,000 results, representing a wealth of algorithm implementations that could benefit from improvement. This is why we believe an opportunity exists to create a subfield focused on generating code from existing optimization algorithms. Building on this, our work explores using LLMs to enhance implemented algorithms, enabling the models to uncover new heuristics overlooked by experts in the original implementations. In this sense, our approach treats LLMs as assistants to researchers, not replacements. By doing so, we leverage LLMs to work with existing code (as context) and improve it using the current knowledge these black-box model models have gained during their pre-training phase.

\subsection{Maximum Independent Set (MIS) Problem} 
\begin{figure}[h]
\centering
\begin{center}
  \includegraphics[width=0.7\linewidth]{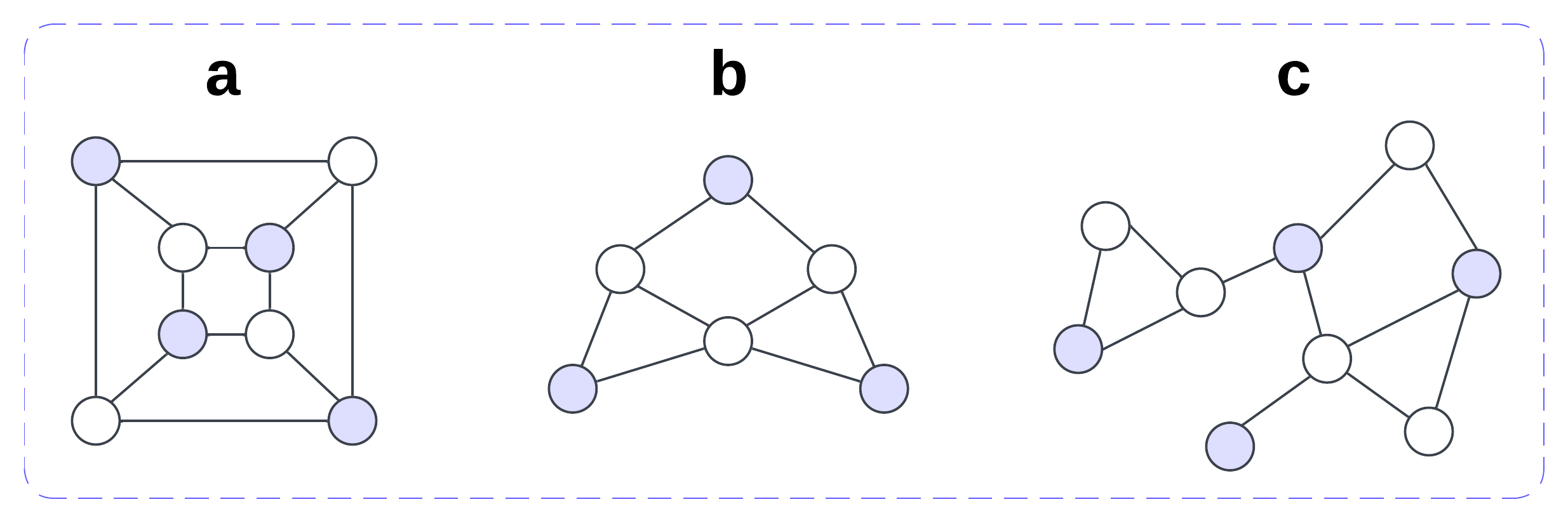}
  \caption{Examples of maximum independent sets.}
  \label{fig:MIS_examples}
\end{center}
\end{figure}

To validate our hypothesis that LLMs can enhance existing optimization algorithms, we will employ an algorithm (detailed in the following subsection) to solve the Maximum Independent Set (MIS) problem, a well-studied and NP-hard combinatorial optimization problem with applications in network design, scheduling, and bioinformatics. Formally, the MIS problem is defined as follows: Given an undirected graph $G = (V, E)$, the objective is to find a largest subset $S \subseteq V$ such that no two vertices in $S$ are adjacent in $G$, i.e., there is no edge $(u, v) \in E$ for all pairs $u \not= v \in S$. Figure~\ref{fig:MIS_examples} shows three examples of optimal MIS solutions (non-white nodes) in different graphs.

\subsection{CMSA}
Construct, Merge, Solve \& Adapt (CMSA) is a hybrid metaheuristic, also known as a matheuristic, that combines elements of classical metaheuristics with exact solvers (such as Integer Linear Programming (ILP) solvers) for combinatorial optimization~\cite{BLUM201675}. Each CMSA iteration is a sequence of four fundamental phases:
\begin{enumerate}
    \item \textbf{Construct:} Generates solutions probabilistically through a probabilistic greedy mechanism (remember this phase; it will be key in the next section).
    
    \item \textbf{Merge:} Combines solution components from the generated solutions to form a reduced subproblem.
    
    \item \textbf{Solve:} Applies an exact solver (in the case of the MIS: an ILP solver) to optimally solve the reduced subproblem.
    
    \item \textbf{Adapt:} Updates parameters and data structures based on the quality of the solution returned by the solver.
\end{enumerate}
This hybrid architecture combines the efficiency of metaheuristics for search space exploration with the precision of exact methods in reduced spaces. The CMSA algorithm is controlled by some key parameters. First, and most importantly, each solution component has an associated \textit{age} value, which is initialized to zero when a component is added to the subproblem. Moreover, a solution component in the subproblem is subject to an increase of its age value, in case it does not form part of the subproblem's solution generated by the exact solver. In this context, the parameter \(age_{max}\) plays a crucial role, defining the maximum age of solution components before they are removed from the incumbent subproblem, thereby preventing stagnation and encouraging diversity. Other parameters include \(n_a\), which sets the number of solution constructions per iteration, \(t_{max}\), the total CPU time, \(t_{limit}\), the time limit for the ILP solver per iteration, and \(0 \leq d_{rate} \leq 1\), which controls the determinism rate for solution construction. Note that higher values of \(d_{rate}\) lead to more deterministic solution constructions, while lower values increase randomness. Well-chosen values for these parameters ensure a balanced exploration of the search space, effective exploitation of promising areas via exact subproblem solving, and efficient use of computational resources.

We selected CMSA for the purpose of this paper due to its relative complexity in implementation as compared to simple metaheuristics. For instance, implementing CMSA for the MIS problem in C++ presents significant technical challenges. The four phases must be precisely defined, properly integrated, and implemented efficiently to leverage the framework's full potential. This complexity would only increase when dealing with more sophisticated optimization problems. In essence, a CMSA implementation demands expertise not only in metaheuristics and exact methods but also proficiency in C++ (or whatever programming language is chosen for implementation).

For our study, we utilize the original C++ implementation to solve the MIS problem, which was provided by CMSA's inventor and can be downloaded from our website (\url{https://imp-opt-algo-llms.surge.sh/}). This choice serves two crucial purposes: first, it ensures we are working with a well-implemented version of the algorithm, and second, it presents an interesting opportunity to test LLMs' capability to enhance expert-written code. This scenario allows us to evaluate whether LLMs can identify potential improvements even in code developed by domain experts. Our findings are presented in the following section.

\section{LLM-Improved CMSA for MIS}\label{sec:integration}

In this section, we introduce a methodology for utilizing LLMs to improve existing optimization algorithms. To present our research and the technical approach in an accessible way, we use the narrative of Dr. Zoe \includegraphics[width=0.018\textwidth]{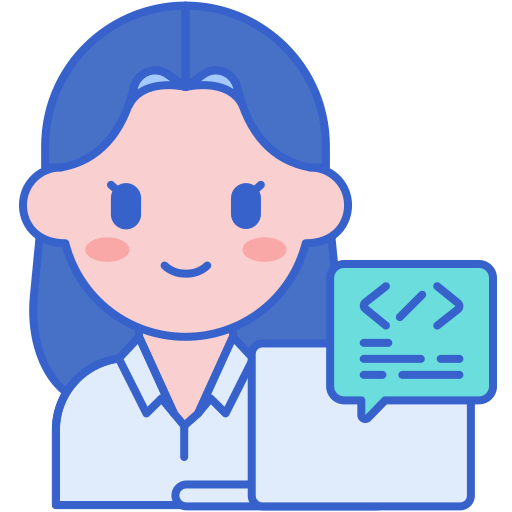}, a fictional senior researcher in hybrid metaheuristics, whose work illustrates how we arrived at this methodology and demonstrates its practical application.

\subsection{Beginning of the story}
Zoe has published extensively on hybrid metaheuristics and has significant expertise in Large Neighborhood Search (LNS) and CMSA, focusing on combinatorial optimization problems such as the Minimum Dominating Set (MDS) and MIS problems. She codes primarily in C++, producing efficient, compact implementations, and only introduces abstractions when necessary, which gives her a competitive edge in developing high-performing algorithms.

Her experience with LLMs is minimal, limited to using ChatGPT to improve her writing or fix package installation issues on Ubuntu. She has not explored LLM-driven code generation and is unfamiliar with the literature in this area. When she hears about LLMs' potential in optimization, she is initially skeptical, aware of their tendency to `hallucinate' from blogs and conversations with colleagues.

\subsection{First Steps with LLMs}
Her perspective begins to shift when a friend suggests using LLMs to update her C++ code to modern standards (e.g., C++23). Intrigued by the possibility of improving her implementation, she decides to experiment with her CMSA code for MIS. \textit{``It is only 400 lines long, nothing too complex,''} she thinks. She crafts a prompt: \textit{"What improvements can I make to my CMSA implementation for solving the Maximum Independent Set (MIS)? \{\{ pastes her 400-line CMSA code here \}\}"}. GPT-4o (which she uses through her free ChatGPT account) provides detailed feedback, with one particular suggestion catching her attention:\footnote{The irrelevant portions of the LLM's responses have been replaced with `…'.}

\begin{tcolorbox}[colback=blue!5, colframe=gray!10!gray,  boxrule=0.2pt, breakable, title=LLM output]
Your implementation of the Construct, Merge, Solver, and Adapt (CMSA) for solving the Maximum Independent Set (MIS) problem appears to be a solid base. However, there are several improvements that could be made to enhance performance, ...

2. Algorithmic Improvements\\
Degree-based heuristic: You're using an increasing degree order (likely vertices with lower degrees are selected first). This is a common heuristic, \textbf{but you might also want to experiment with other heuristics, such as selecting vertices based on a combination of degree and the \underline{age} of vertices (e.g., vertices with low degrees and a low age could be more promising)}. This would add a more dynamic element to your heuristic. ...

\end{tcolorbox}

She is surprised by the model's understanding of CMSA, but especially by its suggestion to improve her heuristic using the \texttt{age} values of solution components (which her original heuristic does not use). To verify her hypothesis, she follows up with another question: \textit{``In my code, what do you understand by \texttt{age}?''} The model responds:

\begin{tcolorbox}[colback=blue!5, colframe=gray!10!gray,  boxrule=0.2pt,breakable, title=LLM output]
...

In the context of CMSA, age is a heuristic mechanism that tracks how long or how frequently vertices have been included in the subproblem. It helps guide the algorithm's decisions about which vertices to keep, prioritize, or exclude from future subproblems. In your code, age is initialized when a vertex is added to the solution for the first time, but to fully utilize the CMSA framework, you should consider incrementing the age of vertices over time and using age as a factor in the selection process. This will allow your CMSA implementation to better explore the solution space and avoid getting stuck in local optima.

...
\end{tcolorbox}

\textit{``Wow!''} Zoe exclaims, impressed by GPT-4o's accurate grasp of CMSA's \texttt{age} concept without explicit explanation. The model's ability to understand the context of her implementation leads her to think beyond just modernizing her code---she sees an opportunity to enhance her MIS heuristic itself.

\subsection{Discovering New Heuristics}

The first thing Zoe does is recall what her heuristic does, as it has been a while since she implemented it. Therefore, she studies function \texttt{generation\_solution()}, which represents the solution construction phase in the CMSA~\cite{blum2024construct}:

\begin{lstlisting}[language=C++,
    caption=Probabilistic greedy algorithm for MIS in CMSA,
    label={lst:original-heuristic-cmsa-1},
    backgroundcolor=\color{backcolour},   
    commentstyle=\color{codegreen},
    keywordstyle=\color{codepurple},
    numberstyle=\tiny\color{codegray},
    stringstyle=\color{codepurple},
    basicstyle=\ttfamily\footnotesize,
    breakatwhitespace=false,         
    breaklines=true,                 
    keepspaces=true,                 
    numbers=left,                   
    showspaces=false,       
    showstringspaces=false,
    showtabs=false,
    frame=single,
    tabsize=2,]
...
while (int(positions.size()) > 0) {
    double dec = standard_distribution(generator);
    int position = 0;
    if (dec <= determinism_rate) position = *(positions.begin());
    else {
        int max = candidate_list_size;
        if (max > int(positions.size())) max = int(positions.size());
        double rnum = standard_distribution(generator);
        int pos = produce_random_integer(max, rnum);
        set<int>::iterator sit2 = positions.begin();
        for (int i = 0; i < pos; ++i) ++sit2;
        position = *sit2;
    }
    greedy_sol.score += 1;
    (greedy_sol.vertices).insert(increasing_degree_order[position]);
   
    if (age[increasing_degree_order[position]] == -1) {
        age[increasing_degree_order[position]] = 0;
    }
    positions.erase(position);
    for (auto sit =neigh[increasing_degree_order[position]].begin(); sit != neigh[increasing_degree_order[position]].end(); ++sit){ 
        positions.erase(position_of[*sit]);
    }
}
\end{lstlisting}

In other words, Zoe's code implements a greedy randomized construction heuristic that selects exactly one vertex $v_i \in \tilde{V}$ at each step (where $\tilde{V}$ is the set of nodes that can feasibly be chosen), until the MIS solution is complete:
\[
    v_i = \begin{cases}
v_{\text{min}} & \text{if } r \leq \alpha \\
v_{\text{random}} \in \text{CL}(k) & \text{otherwise}
\end{cases}
\]

Where:
\begin{itemize}
    \item $v_{\text{min}}$ is the vertex with minimum degree (among the ones from $\tilde{V}$)
    \item $r$ is a random number between $[0,1]$
    \item $\alpha$ is the determinism rate
    \item $\text{CL}(k)$ is a candidate list of size $k$
    \item $v_{\text{random}}$ is a randomly selected vertex from the candidate list
\end{itemize}

The heuristic combines deterministic greedy selection (based on vertex degree) with randomization to create diverse solutions. It selects vertices either by taking the best available vertex (greedy choice) with probability $\alpha$, or by randomly selecting from a restricted candidate list with probability $(1-\alpha)$.

\subsubsection{New heuristic from the LLM} 
\begin{figure*}[!th]
\centering
\rotatebox{90}{%
  \includegraphics[width=0.73\paperheight, height=\paperwidth, keepaspectratio]{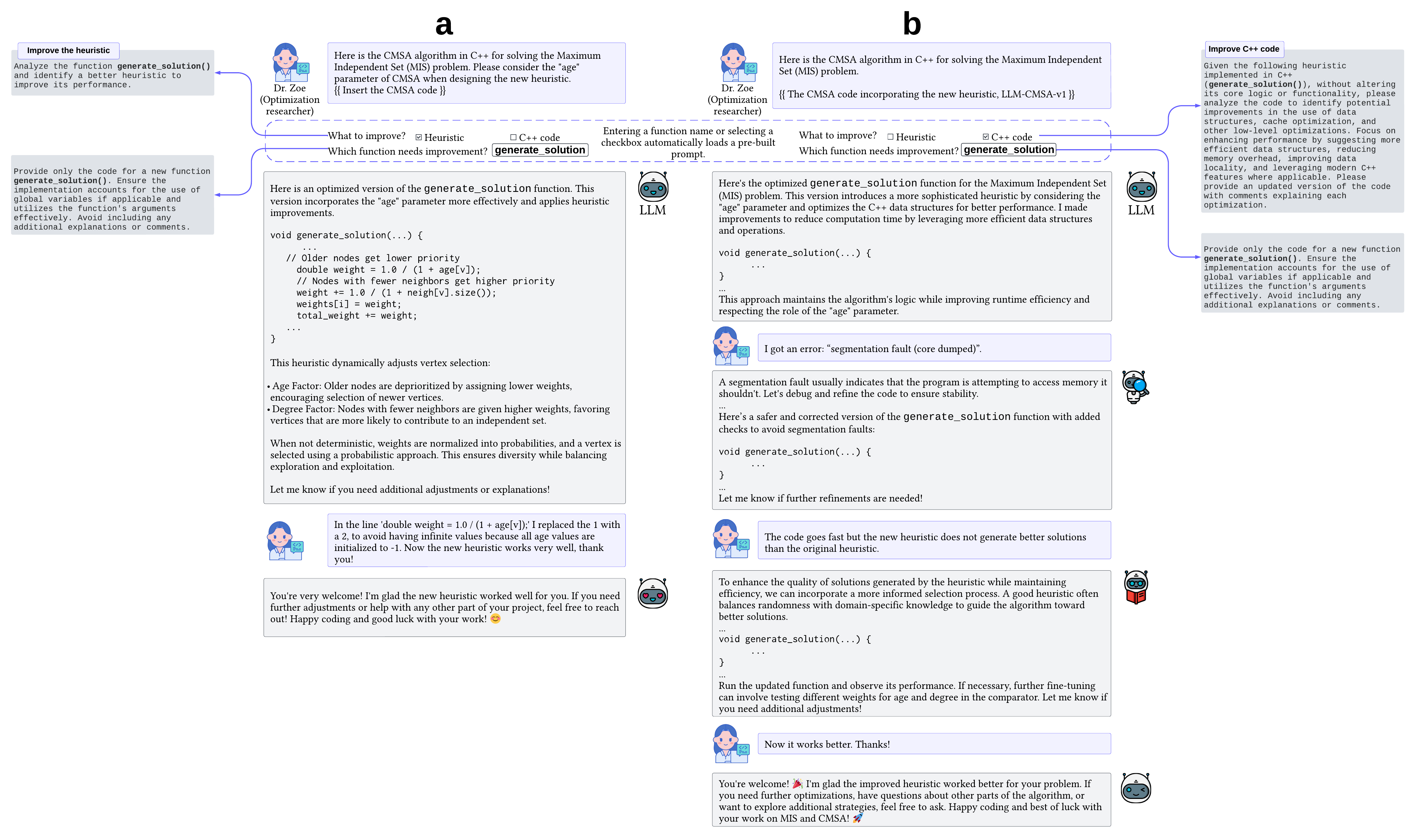}
  }
  \caption{Two LLM interaction patterns: (a) a direct request to improve a heuristic using CMSA's \texttt{age} parameter, and (b) an iterative dialogue to enhance both heuristic quality and C++ performance through error correction. Both use in-context learning as a prompting strategy~\cite{dong2024surveyincontextlearning, Li2023-ij}.}
  \label{fig:chatbot-interaction}

\end{figure*}

To enhance her heuristic, as suggested by the LLM, by incorporating CMSA's \texttt{age} parameter, Zoe provides another prompt to the LLM. In response, the LLM improves the heuristic by introducing a weighted vertex selection mechanism and provides Zoe with the corresponding C++ code. For better context, please refer to case (a) of Figure~\ref{fig:chatbot-interaction} first. In particular, the updated heuristic now considers both node degrees and current age values. The vertex selection mechanism designed by the LLM can technically be described as follows:
\[
v_i = \begin{cases}
    \text{argmin}\{P_w(v_j) \mid v_j \in \tilde{V}\} & \text{if } r \leq \alpha \\
    \text{roulette-wheel selection w.r.t.~$P_w(.)$ values} & \text{otherwise}
\end{cases}
\]
where $P_w(v_j)$ is a weighted probability:
\begin{equation}\label{eq:pw}
P_w(v_j) = \frac{w(v_j)}{\sum_{v_l \in V} w(v_l)}, \quad w(v) = \frac{1}{2 + age(v)} + \frac{1}{1 + \text{degree}(v)}    
\end{equation}
The weight function $w(v)$ favors vertices with low age and degree values to promote diversity in the selection process. Thus, the LLM was able to change the original heuristic by incorporating the age values to diversify the node selection process, while balancing it with degree information through a composite weight function. The more Zoe thinks about it, the better she likes the idea of using the age values in this way. It makes a lot of sense to decrease the probability of solution components with high age values being incorporated in newly constructed solutions. Zoe recognizes that the LLM had to understand the context of CMSA to propose this change. However, Zoe identifies a minor implementation error in the LLM-generated code; see Figure~\ref{fig:chatbot-interaction} (a). She corrects it, noting that adding 1 to \texttt{age} might cause a division by zero, as \texttt{age} values in CMSA are set to -1 for those solution components that do not form part of the subproblem. This highlights that LLMs require feedback~\cite{Chen2023-jk} and cannot be fully autonomous. Finally, we would like to highlight that this use of the \texttt{age} values never occurred to anyone working on CMSA algorithms.


\begin{tcolorbox}[colback=gray!5, colframe=gray!10!gray,  boxrule=0.5pt]
This new CMSA variant is henceforth called \texttt{LLM-CMSA-V1}. It was obtained by replacing the \texttt{generate\_solution()} of the original CMSA implementation with the new function provided by the LLM.
\end{tcolorbox}

\subsubsection{Improving \texttt{LLM-CMSA-V1} with the LLM} After having obtained the new heuristic, Zoe begins to wonder if the LLM could further improve it. Therefore, she inputs the following prompt: \textit{``Are there ways to enhance the dynamic selection heuristic to allow for a more diverse and advanced search?''} The LLM responds with several suggestions, one of which involves incorporating the concept of entropy. The C++ code provided by the LLM is characterized by a corresponding change, which involves replacing the definition of $P_w(.)$ (see Equation~\ref{eq:pw}) with the following entropy-adjusted probabilities:
\[
P_H(v_j) = \frac{P_w(v_j) + H}{\sum_{v_l \in V} P_w(v_l) + H} \enspace,
\]
where
\[
H = -\sum_{v_l \in V} P_w(v_l)\log(P_w(v_l)) \enspace.
\]
The entropy adjustment increases selection diversity by adding the system's uncertainty to each probability.

\begin{tcolorbox}[colback=gray!5, colframe=gray!10!gray,  boxrule=0.5pt]
This CMSA variant is henceforth called \texttt{LLM-CMSA-V2}. We directly replaced the \texttt{generate\_solution()} function of the original CMSA with the LLM-generated code.
\end{tcolorbox}

\subsection{Code Optimization Strategies}
\label{sec:code_optimization}

After discovering the two new versions of CMSA outlined above with the help of the LLM, another idea occurs to Zoe: \textit{``Could the LLM create an improvement at the C++ code level?''} In other words, is there a more efficient way of implementing the new CMSA variants without affecting their logic, by using, for example, more efficient data structures or even low-level techniques? Although Zoe is an expert in C++, she knows that C++ is a very broad language with many features she is unfamiliar with. Perhaps the LLM, with its vast knowledge base, can find a better way to implement the heuristics without altering their logic.  

Zoe creates the following prompt and provides it to the LLM, together with the C++ code of \texttt{LLM-CMSA-V1} and \texttt{LLM-CMSA-v2}:\footnote{This prompt uses the ``C++ code'' checkbox, as shown in Figure~\ref{fig:chatbot-interaction} (b).}

\begin{tcolorbox}[colback=blue!5, colframe=gray!10!gray,  boxrule=0.2pt,breakable, title=Human prompt]
Given the following heuristic implemented in C++ (\texttt{generate\_solution()}), without altering its core logic or functionality, please analyze the code to identify potential \textbf{improvements in the use of data structures, cache optimization, and other low-level optimizations. Focus on enhancing performance by suggesting more efficient data structures, reducing memory overhead, improving data locality, and leveraging modern C++ features where applicable.} Please provide an updated version of the code with comments explaining each optimization.
\end{tcolorbox}

However, she notices that the LLM occasionally generates code with bugs, such as memory management issues in C++ (as shown in Figure~\ref{fig:chatbot-interaction} (b)). She resolves these errors by iteratively copying and pasting the error messages back into the LLM, engaging in a trial-and-error feedback process (a common practice in code generation; see~\cite{Chen2023-jk}). Eventually, the model produces code that compiles successfully. Upon reviewing the final output, she is surprised to find that the code is more complex and harder to read and employs unfamiliar techniques. For example, it replaces the \texttt{set} and \texttt{vector} data structures with more advanced, low-level alternatives:\footnote{The comments in the code were generated by the LLM.}

\begin{lstlisting}[language=C++,
    caption=Fragment code optimization suggested by LLM,
    label={lst:original-heuristic-cmsa-2},
    language=C++,
    backgroundcolor=\color{backcolour},   
    commentstyle=\color{codegreen},
    keywordstyle=\color{codepurple},
    numberstyle=\tiny\color{codegray},
    stringstyle=\color{codepurple},
    basicstyle=\ttfamily\footnotesize,
    breakatwhitespace=false,         
    breaklines=true,                 
    keepspaces=true,                 
    numbers=left,                   
    showspaces=false,       
    showstringspaces=false,
    showtabs=false,
    frame=single,
    tabsize=2,]
#include <bitset>
...
// Constants for optimization
constexpr size_t BLOCK_SIZE = 64;  // CPU cache line size
...
// Bitset for boolean operations
std::bitset<32768> available; 
// Adjust size based on max n_of_vertices  
available.set();
// Aligned vector to optimize cache usage
alignas(BLOCK_SIZE) std::vector<int> active_vertices;
active_vertices.reserve(n_of_vertices);
\end{lstlisting}
Meanwhile, her original code was as shown in the following listing.

\begin{lstlisting}[
    language=C++,
    caption=Fragment of original code from CMSA,
    label={lst:original-heuristic-cmsa-3},
    language=C++,
    backgroundcolor=\color{backcolour},   
    commentstyle=\color{codegreen},
    keywordstyle=\color{codepurple},
    numberstyle=\tiny\color{codegray},
    stringstyle=\color{codepurple},
    basicstyle=\ttfamily\footnotesize,
    breakatwhitespace=false,         
    breaklines=true,                 
    keepspaces=true,                 
    numbers=left,                   
    showspaces=false,       
    showstringspaces=false,
    showtabs=false,
    frame=single,
    tabsize=2,]
set<int> positions;
vector<int> position_of(n_of_vertices, 0);
for (int i = 0; i < n_of_vertices; ++i) {
    positions.insert(i);
    position_of[increasing_degree_order[i]] = i;
}
\end{lstlisting}

The logic remains the same, except for changes in variable names: \texttt{positions} is replaced with \texttt{available,} and \texttt{position\_of} is substituted with \texttt{active\_vertices.}\footnote{Explicitly instructing the prompt to retain the variable names might have avoided this issue.} 

Zoe is not entirely convinced that these changes (among others) will result in any tangible benefit. Nevertheless, the code compiles and executes correctly.\footnote{Visit our project website for more details on the C++ optimizations suggested by GPT-4o.}

\begin{tcolorbox}[colback=gray!5, colframe=gray!10!gray, boxrule=0.5pt]
The two new CMSA variants with these performance improvements will be named: \texttt{LLM-CMSA-V1-PERF} and \texttt{LLM-CMSA-V2-PERF}.
\end{tcolorbox}

Zoe's story highlights the opportunities for optimization researchers to improve their algorithms and implementations, fostering creativity and leveraging the LLM as a versatile assistant.

\begin{figure*}[!h]
    \centering
    \begin{subfigure}{0.35\textwidth}
        \centering
        \includegraphics[width=\textwidth]{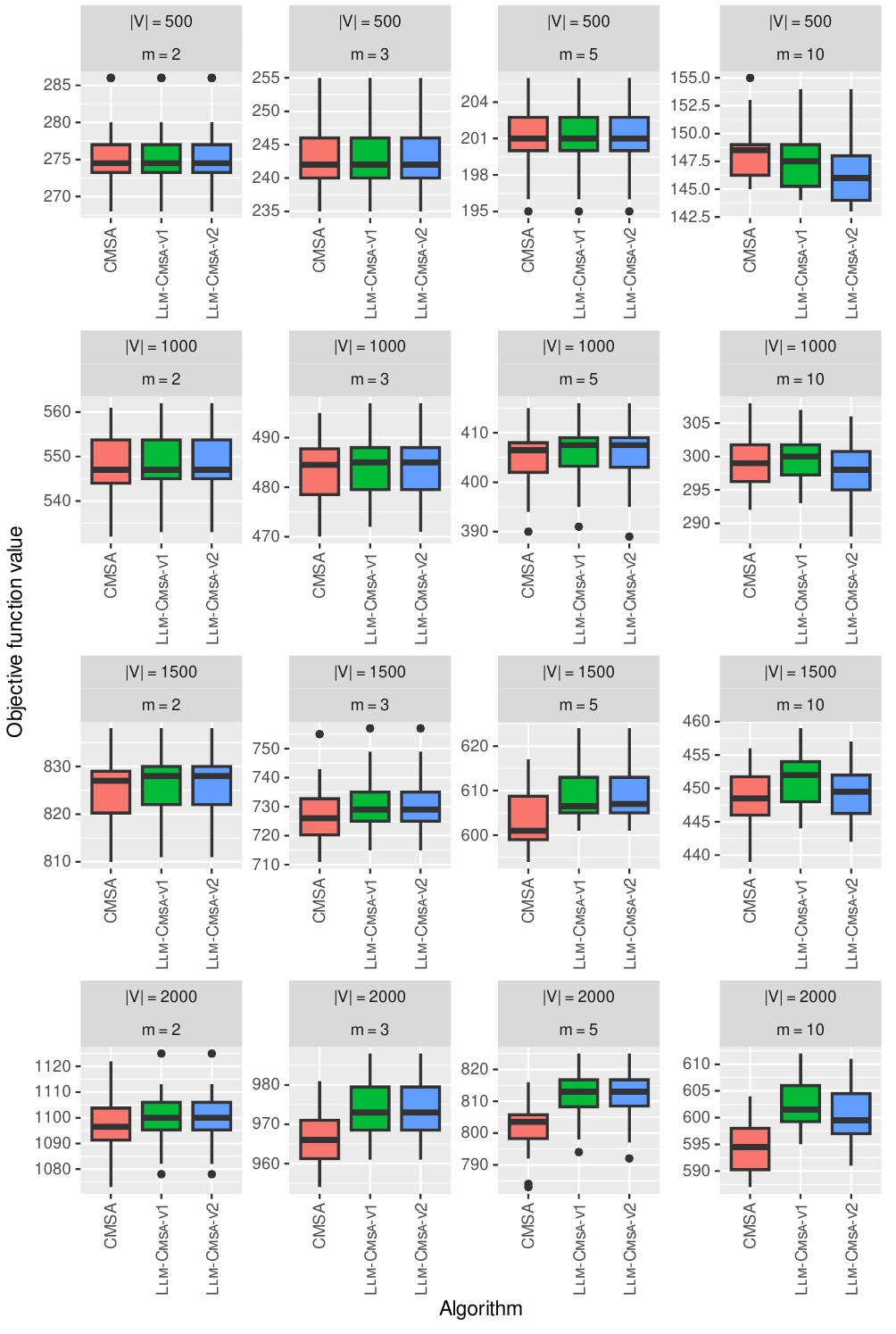}
        \caption{Barabási–Albert}
        \label{fig:barabasi}
    \end{subfigure}
    \hfill
    \begin{subfigure}{0.35\textwidth}
        \centering
        \includegraphics[width=\textwidth]{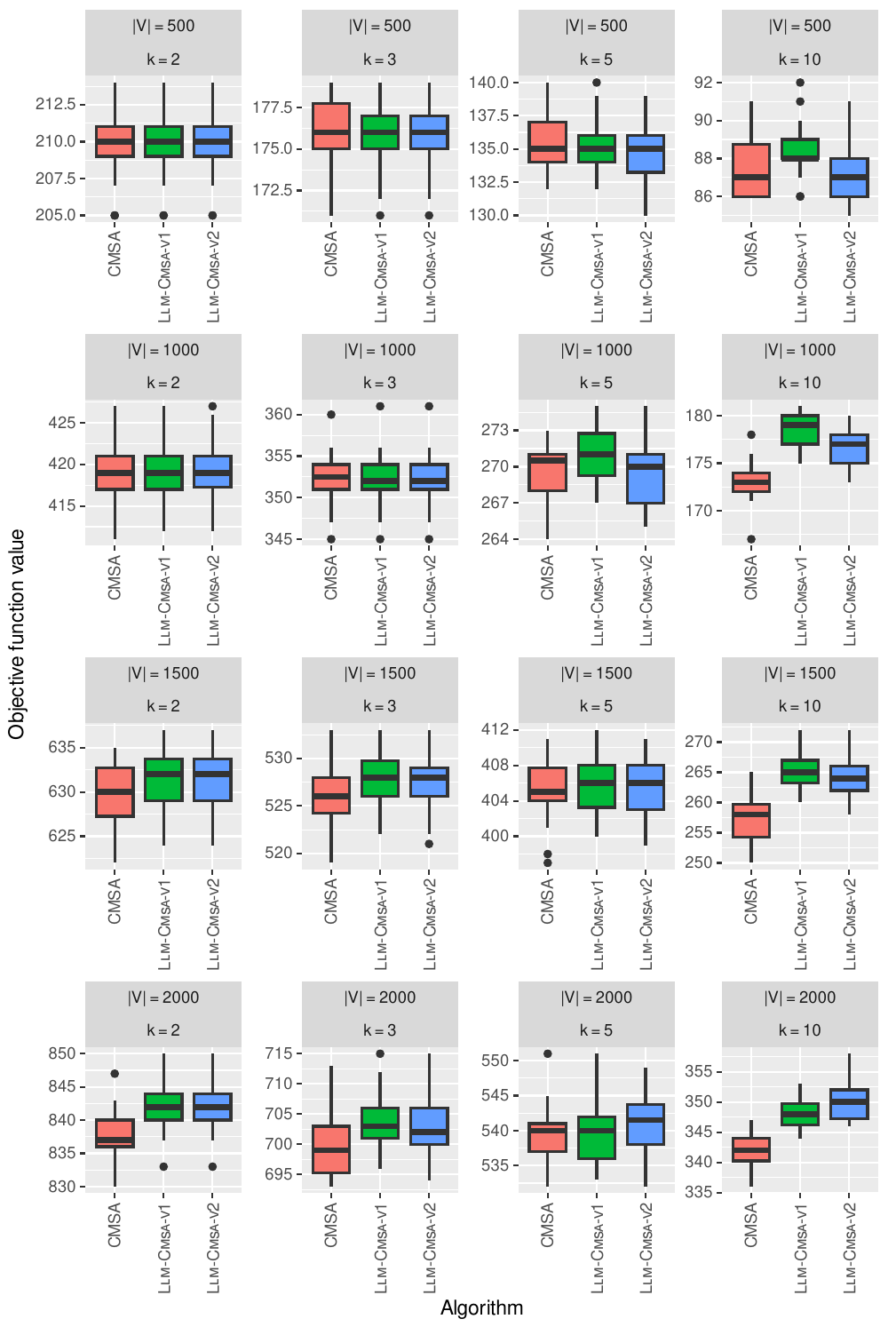}
        \caption{Watts–Strogatz}
        \label{fig:watts}
    \end{subfigure}
    \hfill
    \begin{subfigure}{0.35\textwidth}
        \centering
        \includegraphics[width=\textwidth]{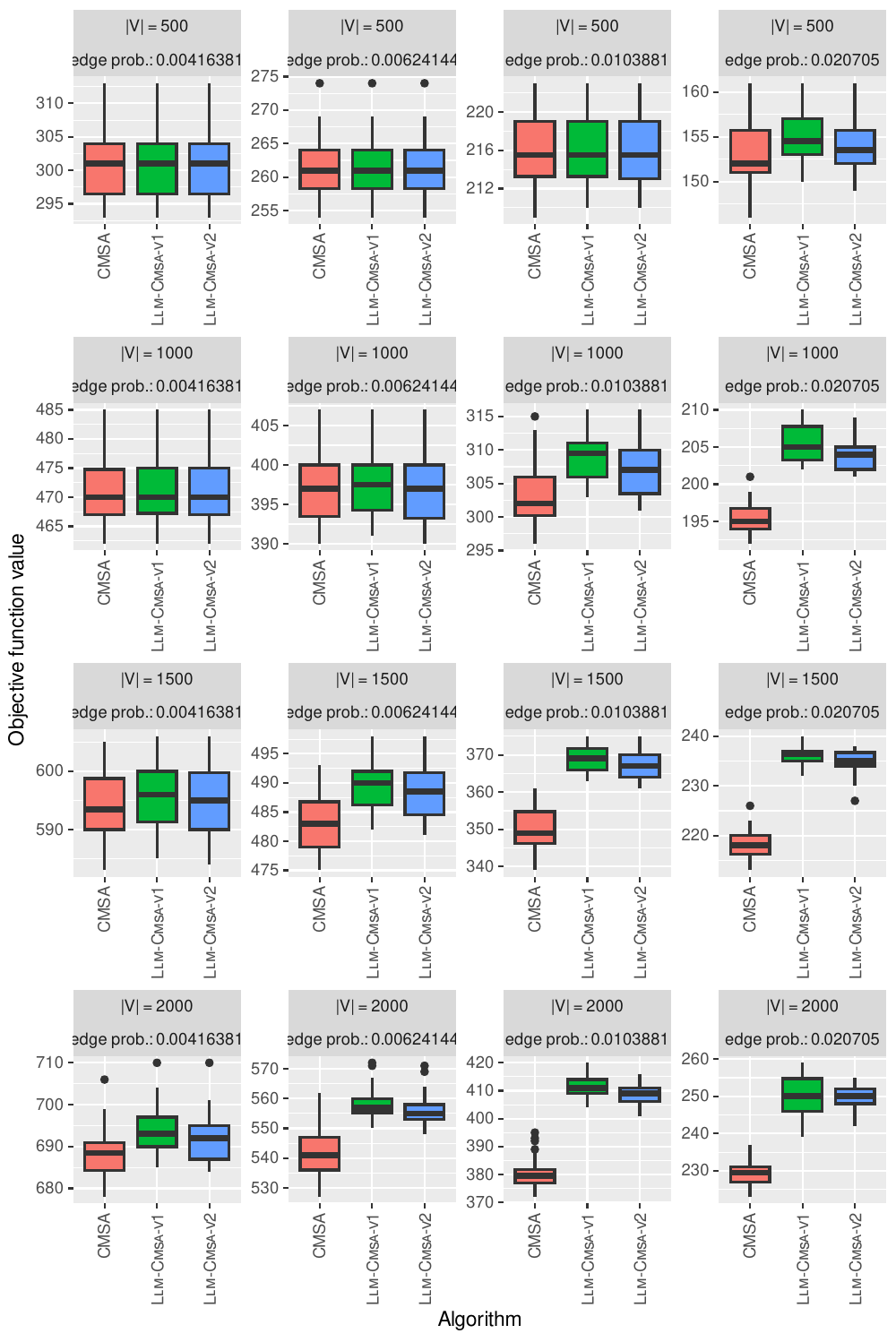}
        \caption{Erdős–Rényi }
        \label{fig:erdos}
    \end{subfigure}
    \caption{Comparative analysis of solution quality: Original CMSA vs. LLM-CMSA variants (V1 and V2).}
    \label{fig:results-graphs}
\end{figure*}

\subsection{Reproducility} Although it is not possible to replicate the exact results of an LLM, due to their autoregressive nature that predicts the most probable token based on a probability distribution (with the next token being determined stochastically)~\cite{kamath2024large}, it is possible to reproduce similar responses by using the same prompts, the same LLM, and its parameters. For this reason, our repository (\url{https://imp-opt-algo-llms.surge.sh/}) includes a \textit{chatbot} that implements the same prompts used in our research (as shown in Figure~\ref{fig:chatbot-interaction}). In fact, each element in Figure~\ref{fig:chatbot-interaction} (textbox and checkbox) loads pre-built prompts, known as in-context prompts~\cite{dong2024surveyincontextlearning, Li2023-ij}, to eliminate the need for manual input. Thus, our chatbot features two types of in-context prompts: (1) external ones, related to the C++ CMSA code for the MIS, and (2) internal ones, focused on improving the heuristic, the C++ code, and specifying which function in the code requires enhancement. Next, we will assess the quality of the heuristics proposed by the LLM.

\begin{figure*}[!h]
    \centering
    \begin{subfigure}{0.6\textwidth}
        \centering
        \includegraphics[width=\textwidth]{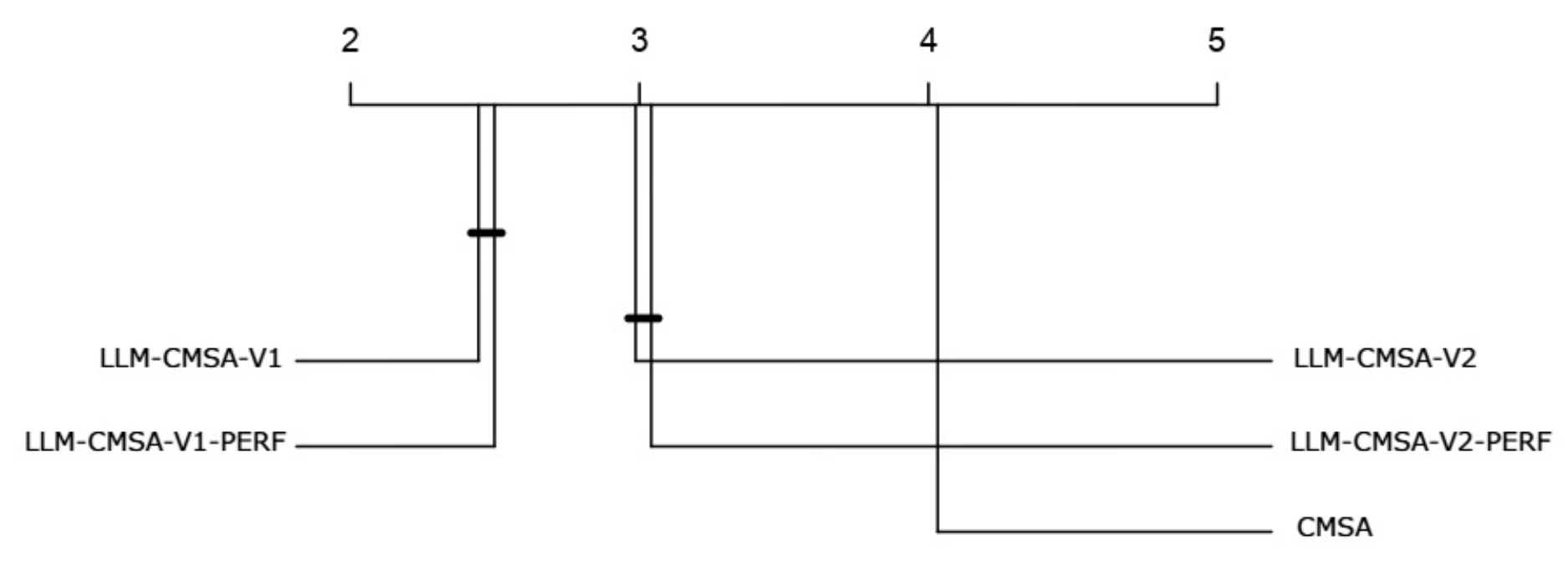}
        \caption{Barabási–Albert}
        \label{fig:diff-barabasi}
    \end{subfigure}
    \hfill
    
    \begin{subfigure}{0.6\textwidth}
        \centering
        \includegraphics[width=\textwidth]{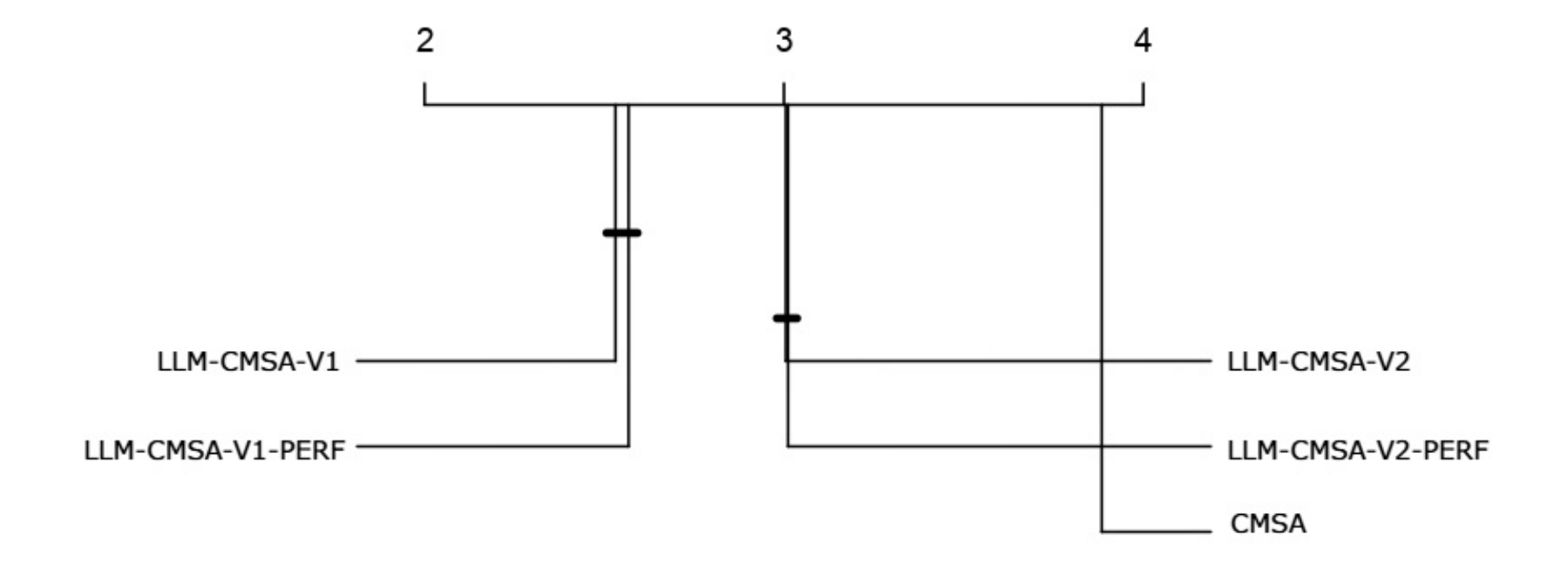}
        \caption{Watts–Strogatz}
        \label{fig:diff-watts}
    \end{subfigure}
    \hfill
    
    \begin{subfigure}{0.6\textwidth}
        \centering
        \includegraphics[width=\textwidth]{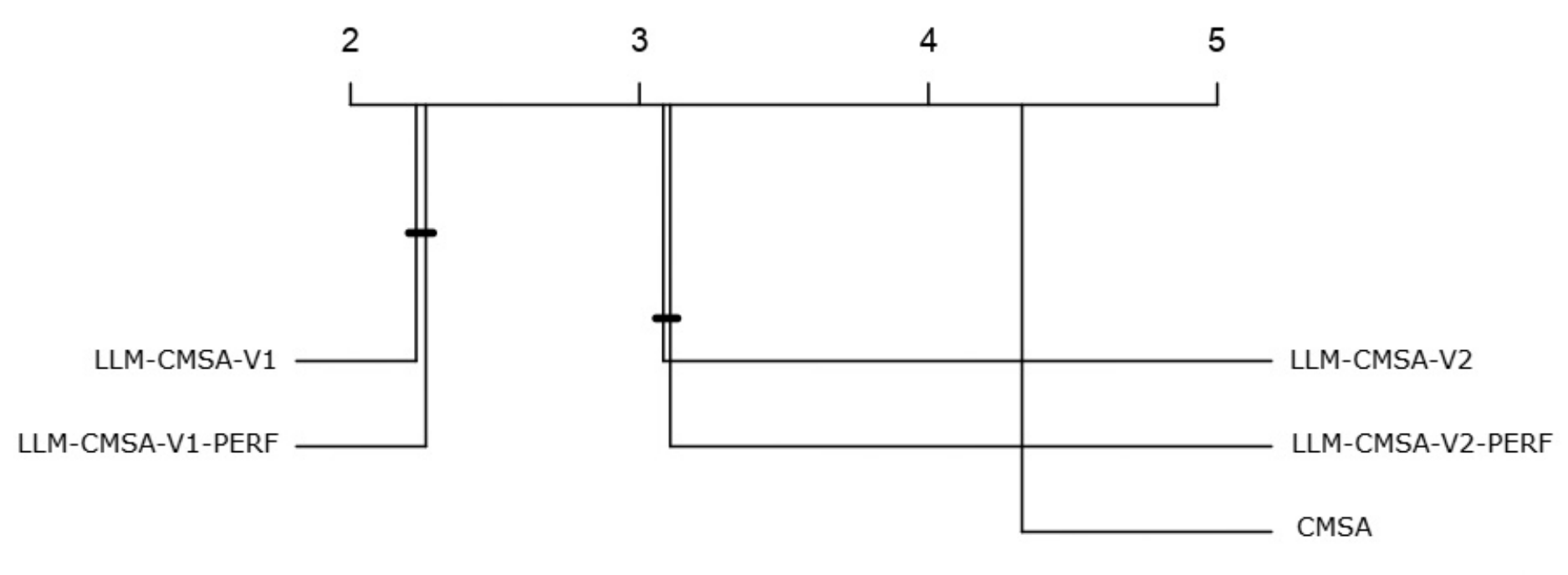}
        \caption{Erdős–Rényi}
        \label{fig:diff-erdos}
    \end{subfigure}
    
    \caption{Critical difference (CD) plots for all graph types}
    \label{fig:diff-general}
\end{figure*}

\begin{figure}[htbp]
    \centering
    
    \begin{subfigure}{0.6\linewidth}
        \centering
        \includegraphics[width=\textwidth]{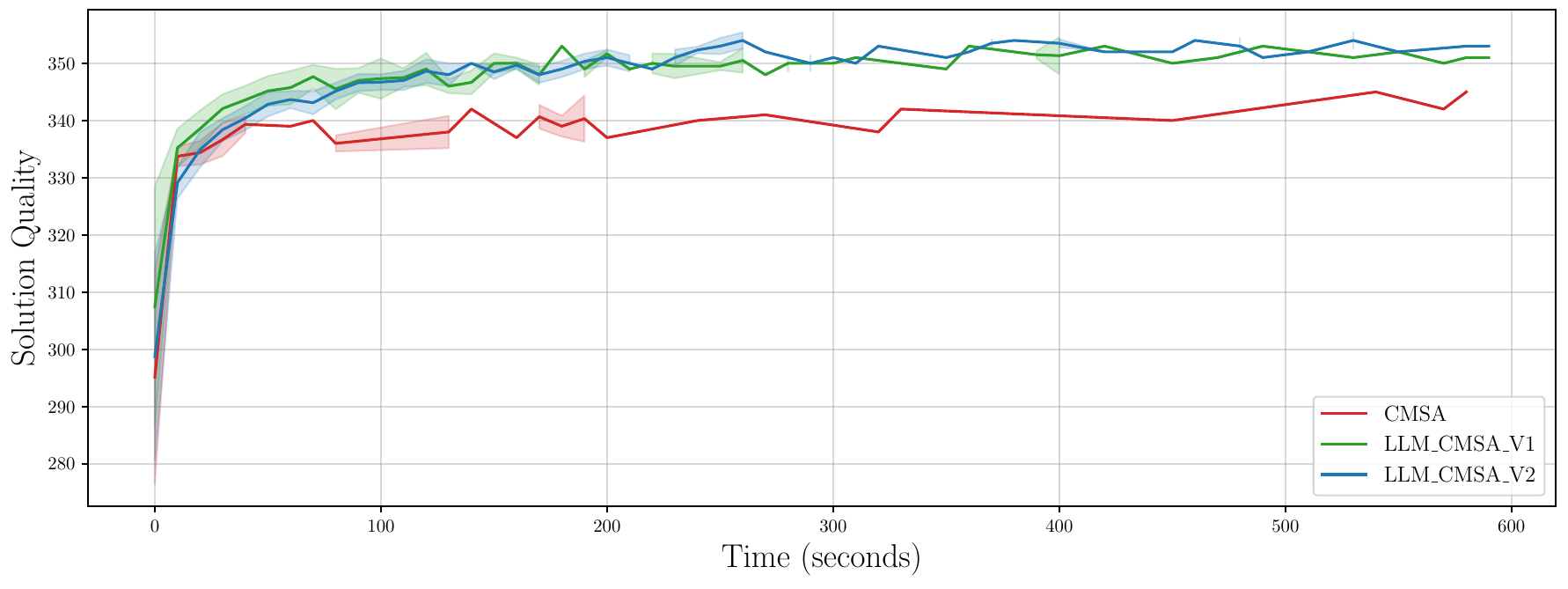}
        \caption{Watts–Strogatz (n2000, k10)}
        \label{fig:convergece-watts}
    \end{subfigure}
    
    \begin{subfigure}{0.6\linewidth}
        \centering
        \includegraphics[width=\textwidth]{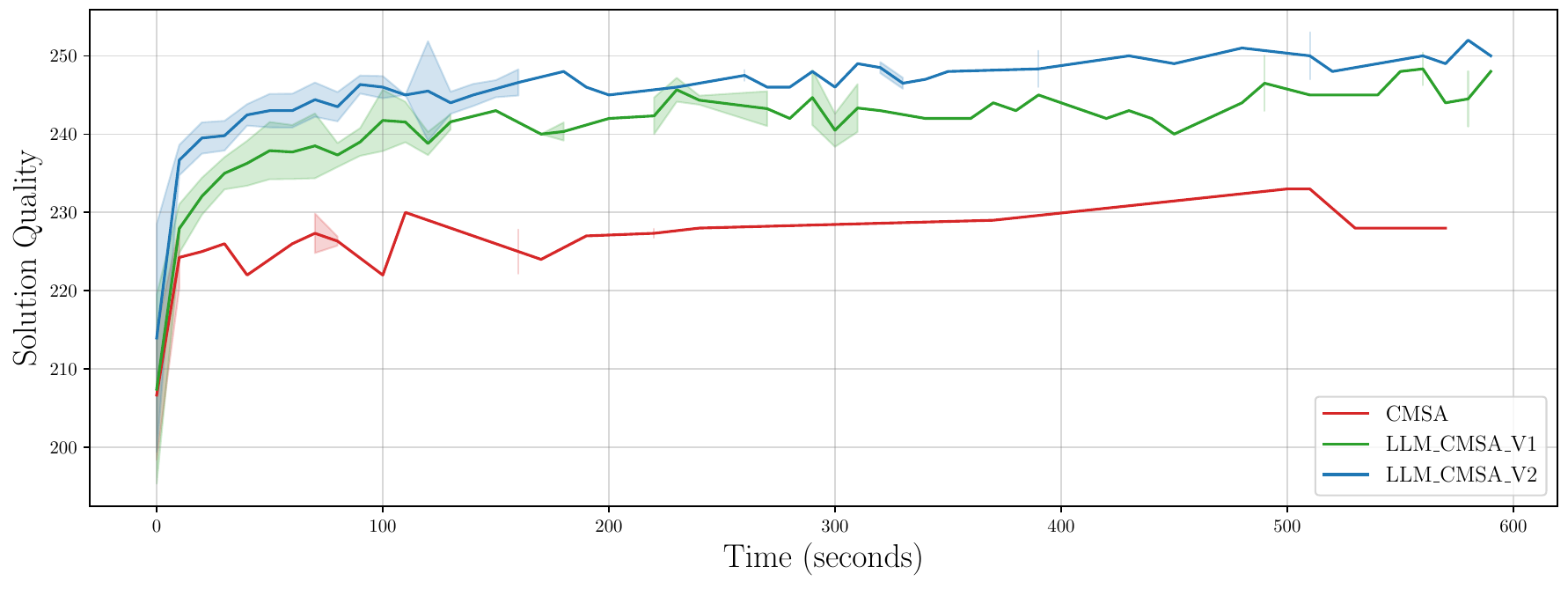}
        \caption{Erdős–Rényi (n2000, 020705)}
        \label{fig:convergence-erdos}
    \end{subfigure}
    
    \caption{Examples of algorithm evolution over time}
    \label{fig:convergence-general}
\end{figure}

\section{Empirical Evaluation}\label{sec:evaluation}

This section is divided into two parts: the preliminary phase (setup, benchmark, and CMSA parameter tuning) and the results.

\subsection{Preliminary}

First, we used Chatbot Arena~\cite{chiang2024chatbotarenaopenplatform} to test various LLMs without incurring costs.~\footnote{\url{https://lmarena.ai/}} For our experiments, we finally selected GPT-4o (version: 2024-11-20)\footnote{The parameters used for the LLM were the default values: \texttt{temperature = 0.7} (controls randomness in responses), \texttt{top-p = 1} (nucleus sampling threshold for token probability), and \texttt{max-output-tokens = 2048} (maximum number of tokens in the output).} as it is one of the top-performing models to date. Experiments concerning algorithm variants \texttt{CMSA}, \texttt{LLM-CMSA-V1}, and \texttt{LLM-CMSA-V2} were conducted on a cluster equipped with Intel® Xeon® CPU 5670 processors (12 cores at 2.933 GHz) and 32 GB of RAM.


Our benchmark set consists of three types of graphs: Barabási-Albert, Watts-Strogatz, and Erdős-Rényi graphs, with four different sizes and four density levels (see Figure~\ref{fig:results-graphs}). For each combination of size and density level, the benchmark set contains 1 tuning instance and 30 testing instances. This makes a total of 48 tuning instances and 1440 testing instances. The parameters of all three CMSA variants (original, V1, V2) were tuned using \texttt{irace}, a tool for tuning algorithm parameters based on problem instances~\cite{irace}. Due to the lack of space (and not being the important point of this paper), we do not report the final values here. We used 150, 300, 450, and 600 CPU seconds as computation time limits for graphs of the four different sizes.

\subsection{Numerical Results}

In addition to the three tuned algorithms, we also tested two variants obtained by using the efficiency-improved C++ codes described in Section~\ref{sec:code_optimization}. These two variants are henceforth called \texttt{LLM-CMSA-V1-PERF} and \texttt{LLM-CMSA-V2-PERF}. Each of the five algorithms was applied exactly once to each testing instance. The results of the main three variants are reported employing box plots in Figure~\ref{fig:results-graphs}. The considered graph size and density level are indicated at the top of each box plot. The main key observation is that both LLM-generated CMSA variants outperform the standard CMSA variant with growing graph size and density. This clearly shows that the suggestion of the LLM to make use of the age values for solution component selection during solution construction was a very good one.




To be able to make statistical claims, we produced so-called critical difference (CD) plots (see Figure~\ref{fig:diff-general}), in which the algorithms' whiskers show their average ranking concerning a set of problem instances. Moreover, algorithm whiskers are connected by a bold horizontal bar in case they perform statistically equivalent. Figure~\ref{fig:diff-general} shows three CD plots, one for each graph type. In these plots we also included the efficiency-optimized LLM-generated CMSA variants. The following observations can be made. Even though both \texttt{LLM-CMSA-V1} and \texttt{LLM-CMSA-V2} outperform \texttt{CMSA} with statistical significance for all three graph types, in all three cases the first variant outperforms the second variant. This means that the idea of using the entropy of the selection probabilities was not fruitful. Moreover, the efficiently-optimized algorithm variants are statistically equivalent to their non-optimized counterparts. This means that, even though they might save RAM, which is a non-tested hypothesis, they do not benefit by producing better results. 


Finally, Figure~\ref{fig:convergence-general} shows two representative examples of the convergence behavior based on 10 runs of the main three CMSA variants. In both examples, the LLM-generated CMSA variants quickly produce solutions of a quality that standard CMSA does not even obtain at the end of its runs. 





\section{Discussion}\label{sec:disc}

Our study has the following limitations: due to space constraints, we tested only one LLM (GPT-4o) for CMSA improvements, and future work could benefit from comparing additional LLMs, especially open-weight ones. Additionally, we did not explore using GPT-4o to enhance other complex algorithms for additional optimization problems. These limitations could be addressed in an extended article. Despite these limitations, our comprehensive analysis of CMSA with new heuristics for solving MIS instances presents promising opportunities for new lines of research emerging from our work:

\begin{enumerate} \item \textbf{Creation of specialized benchmarks.} Our experiments focused on a single optimization algorithm (CMSA) applied to one problem (MIS). Just as general-purpose code generation has well-established benchmarks, there is a clear need for specialized benchmarks tailored to optimization. This would enable us to assess which models can effectively `discover' better heuristics for existing algorithms.

\item \textbf{Integration of LLM-based agents.} Figure~\ref{fig:chatbot-interaction} shows a manual interaction with an LLM. However, tasks such as executing code and correcting errors (feedback process) could be delegated to autonomous code agents capable of communicating with each other. Developing a platform where these agents can collaborate on \textit{existing} optimization algorithms and propose better implementation strategies could lead to significant advancements in the field (see~\cite{Jiang2024-xm, zhang-etal-2024-codeagent}). Some progress has been made in creating a platform that unifies the design of optimization algorithms from scratch (e.g., \textsc{LLM4AD}~\cite{liu2024llm4adplatformalgorithmdesign}).

\item \textbf{Programming language translation for optimization algorithms.} To enhance maintainability and performance, algorithms may need translation across languages. LLMs can efficiently optimize readability and performance~\cite{Pan2023-hg, Lachaux2020-to}, but specialized LLMs (possibly fine-tuned) are required to ensure coherent translations tailored to optimization algorithms.

\end{enumerate}

An essential question arises: should the LLM receive credit for discovering a superior heuristic compared to a human expert's best implementation? While prompt design certainly influences outcomes, it is worth thinking about questions concerning authorship, ownership, and AI's role in scientific discovery (see~\cite{RePEc:nat:nature:v:620:y:2023:i:7972:d:10.1038_s41586-023-06221-2}).

\section{Conclusions}\label{sec:conclusion}

In our research, we demonstrate that LLMs can be effectively applied to enhance existing optimization algorithms. We used the non-trivial Construct, Merge, Solve, and Adapt (CMSA) algorithm implemented in C++ to solve the classical Maximum Independent Set problem. By leveraging in-context prompts with GPT-4o, the model successfully captured the context of the CMSA implementation and proposed new heuristics for the probabilistic construction phase of CMSA. After conducting a thorough comparative analysis, the suggested heuristics outperformed those implemented by an expert in CMSA.

\section*{Acknowledgments}
C.~Chacón Sartori and C.~Blum were supported by grant PID2022-136787NB-I00 funded by MCIN/AEI/10.13039/501100011033. Moreover, thanks to FreePik, from where we extracted the icons used in Figures~\ref{fig:teaser} and~\ref{fig:chatbot-interaction}.


\end{document}